\documentclass[final]{cvpr}

\usepackage{times}
\usepackage{epsfig}
\usepackage{graphicx}
\usepackage{amsmath}
\usepackage{amssymb}

\usepackage{algorithm}
\usepackage{algpseudocode}
\usepackage{subcaption}
\usepackage{mathtools}
\usepackage{float}
\usepackage{multirow}
\usepackage{color}
\usepackage{wrapfig}
\usepackage{caption}
\usepackage[dvipsnames]{xcolor}
\usepackage{colortbl}
\usepackage{booktabs}
\usepackage{xcolor}
\usepackage{diagbox}

\usepackage{pdfrender}
\newcommand*{\boldcheckmark}{%
	\textpdfrender{
		TextRenderingMode=FillStroke,
		LineWidth=.5pt, %
	}{\checkmark}%
}

\usepackage{array}
\newcolumntype{L}[1]{>{\raggedright\let\newline\\\arraybackslash\hspace{0pt}}m{#1}}
\newcolumntype{C}[1]{>{\centering\let\newline\\\arraybackslash\hspace{0pt}}m{#1}}
\newcolumntype{R}[1]{>{\raggedleft\let\newline\\\arraybackslash\hspace{0pt}}m{#1}}

\newcommand{\adv}{\mathrm{adv}}

\newcommand{\bt}[1]{\textbf{\textit{#1}}}

\newcommand{\sbd}[1]{\small{(\textbf{#1})}}
\newlength\savewidth\newcommand\shline{\noalign{\global\savewidth\arrayrulewidth
		\global\arrayrulewidth 1pt}\hline\noalign{\global\arrayrulewidth\savewidth}}

\definecolor{grey}{rgb}{0.9,0.9,0.9}

\usepackage[pagebackref=true,breaklinks=true,colorlinks,bookmarks=false]{hyperref}

\def\cvprPaperID{4790} %
\def\confYear{CVPR 2021}

\def \submission {} %
\ifx \submission \undefined
	\newcommand{\jk}[1]{\textcolor{blue}{#1}}
	\newcommand{\siva}[1]{\textcolor{brown}{Siva: #1}}
	\newcommand{\sm}[1]{\textcolor{brown}{#1}}
	\newcommand{\raquel}[1]{\textcolor{red}{Raquel: #1}}
\else
	\newcommand{\jk}[1]{{#1}}
	\newcommand{\siva}[1]{}
	\newcommand{\sm}[1]{{#1}}
	\newcommand{\raquel}[1]{}
\fi

\begin{document}

\title{AdvSim: Generating Safety-Critical Scenarios for Self-Driving Vehicles
} 

\author{
Jingkang Wang$^{1,2}$\quad Ava Pun$^{3}$\quad James Tu$^{1,2}$\quad Sivabalan Manivasagam$^{1,2}$\quad Abbas Sadat$^{2}$ \\ Sergio Casas$^{1,2}$\quad Mengye Ren$^{1,2}$\quad Raquel Urtasun$^{1,2}$ \\
{University of Toronto$^{1}$, Uber ATG$^{2}$, University of Waterloo$^{3}$} \\
{\tt\small {\{wangjk, manivasagam, sergio, mren, urtasun\}}@cs.toronto.edu}\ \ \\
 \tt\small{a5pun@uwaterloo.ca} \ \ 
 \tt\small{james.tu@mail.utoronto.ca} \ \ abbas.sadat@gmail.com
}

\maketitle

\begin{abstract}

As self-driving systems become better, simulating scenarios where the autonomy stack may fail becomes more important.
Traditionally, those scenarios are generated for a few scenes with respect to the planning module that takes ground-truth actor states as input.
This does not scale and cannot identify all possible autonomy failures, such as perception failures due to occlusion.
In this paper, we propose AdvSim, an adversarial 
framework to generate safety-critical scenarios for any LiDAR-based autonomy system.
Given an initial traffic scenario, AdvSim modifies the actors' trajectories in a physically plausible manner and updates the LiDAR sensor data to match the perturbed world. 
Importantly, by simulating directly from sensor data, we obtain adversarial scenarios that are safety-critical for the full autonomy stack.
Our experiments show that our approach is general and can identify thousands of semantically meaningful safety-critical scenarios for a wide range of modern self-driving systems. 
Furthermore, we show that the robustness and safety of these systems can be further improved by training them with scenarios generated by AdvSim.

\end{abstract}
\section{Introduction}

Self-driving vehicles (SDV) 
are safety critical applications in which the comprehensive testing is necessary before real-world deployment.
As the performance of self-driving systems becomes better on natural and well-behaved scenarios, 
it becomes of key importance to find scenarios where the system is likely to fail. However, exhaustively searching over all possible scenarios to identify safety critical ones \jk{is} computationally unfeasible,
as there are exponentially many scenario variations due to the combinatorial number of possible lane topologies, actor configurations, trajectories, velocity profiles,  appearance of actors and background, etc. 

Conventional practice in industry for comprehensive testing is a semi-autonomic process that relies on human expertise to create an initial scenario set, where each scenario contains
at most 1 or 2
``actors of interest" (e.g., vehicles 
that interact with the SDV's planned path) with specified initial locations and trajectories~\cite{rosero2020software,madrigal_2018}. 
Scenario variations are then programmatically created by varying the actors' locations and velocity profiles.  
While such scenarios \jk{are valuable}, they only evaluate simple interactions with the SDV and do not test
complex multi-actor interactions, such as lane-merging and unprotected left-turns in dense traffic scenes.
They do not test the autonomy system on \sm{the wide variety}
of scenarios that the SDV may encounter. %
Moreover, human involvement makes this process time-consuming and difficult to scale. 
Manual design may also result in missing testing configurations that 
identify unexpected failure modes, as it is difficult to assess coverage. 
\begin{figure}[t]
	\centering
	\includegraphics[width=\linewidth]{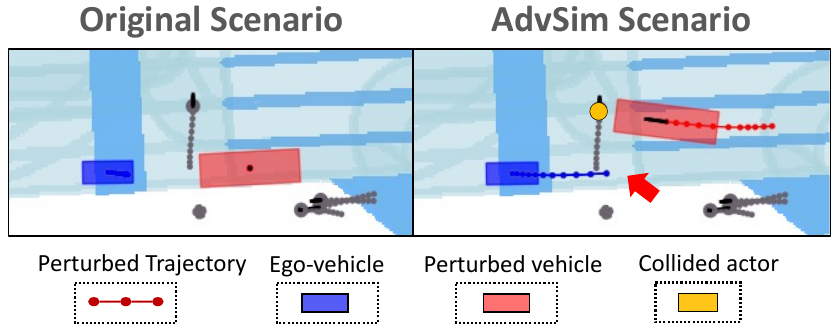}
	\vspace{-0.2in}
	\caption{
		AdvSim modifies traffic scenarios
in a physically plausible manner to produce
autonomy system failure. %
		Here the SDV collides with a crossing pedestrian
		after the behavior of a nearby bus is modified by AdvSim. The collision is indicated by the red arrow.
	}
	\label{fig:demo}
	\vspace{-0.15in}
\end{figure}

\begin{figure*}[t]
	\centering
	\includegraphics[width=\textwidth]{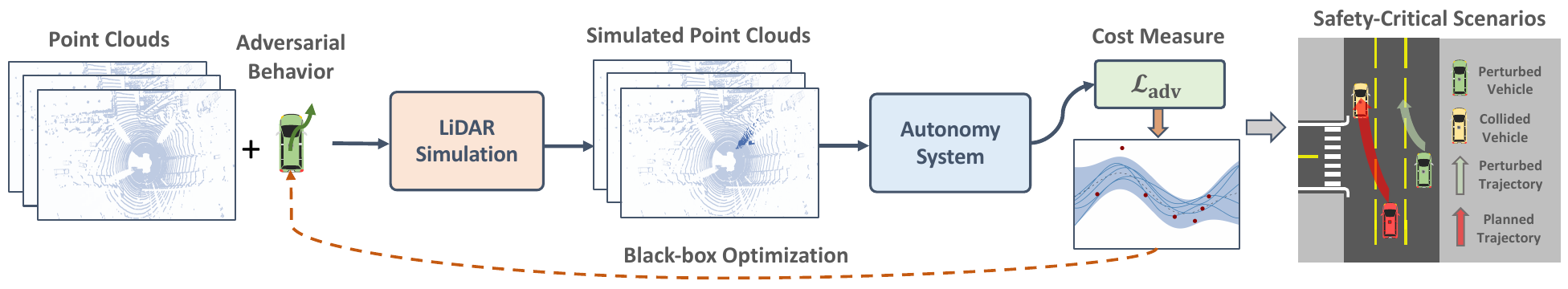}
	\vspace{-5mm}
	\caption{
		\textbf{Overview of our proposed adversarial
scenario generation pipeline.} 
Our goal is to perturb
the maneuvers of interactive actors in an existing scenario with adversarial behaviors that cause realistic autonomy system failures. Given an existing scenario and its original sensor data, we perturb the scenario and update accordingly how the SDV would observe the LiDAR sensor data based on the new scene configuration. We then evaluate the autonomy system on the modified scenario, compute an adversarial objective, and update the proposed perturbation using a search algorithm.
}
	\label{fig:pipeline}
\end{figure*}

\vspace{-4mm}
To address this problem, recent works aim to automate the scenario generation procedure by searching over the set of possible scenarios and identifying high-risk ones according to a specified cost function. 
Most previous works \cite{adv2020learn2collide, adv19failure, adv19falsification, adv2020lane, ding2020multimodal} only consider a motion planning module that has access to the ground-truth state of the actors in the scene.
This overlooks the fact that many adversarial scenarios often involve actors that are hard to identify due to occlusion, or that have trajectory plans that can be difficult to localize and forecast. 
Such issues in the perception and motion forecasting modules of the autonomy system can generate compounding errors that ultimately cause planning failures.
While \cite{o2018scalable, norden2019efficient, adv19bo} test end-to-end image-based self-driving systems, the adversarial scenarios they generate are either generated at a small scale \cite{adv19bo} %
or with respect to simplified imitation learning models that do not reflect the autonomy system of modern self-driving vehicles~\cite{o2018scalable, norden2019efficient}. 
Additionally, most past works either modify the scenario by changing high-level actor behavior \cite{adv2020learn2collide, ding2020multimodal, adv19evolutionary, o2018scalable, norden2019efficient, adv19bo}, or create physically unrealizable trajectories \cite{adv19failure}. This does not allow for physically-plausible fine-grained control of the actor trajectory, such as creating a nudging actor.

In contrast, we want to find
complex  and realistic
safety-critical scenarios at scale for the full 
self-driving system. 
Towards this goal, we  frame the generation of worst-case scenarios as a black box adversarial attack that can test any LiDAR-based autonomy system. 
We explore adversarial perturbations with respect to physically feasible changes in actor behavior, 
since such perturbations provide insight
into the different types of driving situations that are challenging.
This contrasts previous works on black box attacks for perception systems  \cite{meshadv,liu2018beyond,zeng2019adv, james2020adv, advtshirt} that perturb appearance and texture,
\sm{but do not perturb actor behavior.} %

In this paper, we leverage real world traffic scenarios available in standard self-driving datasets and optimize the actors' trajectories jointly to increase the risk of an autonomy system failure.
As our perturbation modifies the \sm{actors' trajectories,}
we need to adjust the sensor data to accurately reflect the actors' new locations.
We therefore adopt a high-fidelity LiDAR simulator~\cite{siva2020lidarsim} that modifies the sensor data accordingly taking into account occlusions. 
After running the black-box autonomy system with modified sensor data as input, we obtain the planned trajectory and 
evaluate how adversarial the scenario was.
Our adversarial objective captures multiple safety factors such as collisions, violations in traffic rules, and uncomfortable driving behaviors. 
We demonstrate the flexibility and scalabity of our approach by generating over 4000 adversarial scenarios for a wide range of modern autonomy systems.
Finally, we leverage AdvSim-generated safety-critical scenarios in training and further improve the safety of autonomy systems.

\section{Related Work}
We first 
give an overview of the development of modern autonomy stack. We then discuss the existing works on safety-critical scenario generation for the planning module.
We also review works
producing 
physically realizable adversarial examples for self-driving perception systems.

\vspace{-3mm}
\paragraph{Self-Driving System:}
\sm{Industry typically decomposes} autonomy systems into 
three \sm{sequential} subtasks:
object detection (perception), motion forecasting (prediction), and planning.
However, these components are developed separately, \sm{and cannot correct}
compounding errors.
Another 
approach is end-to-end self-driving, which \sm{traces}
back to the seminal work ALVINN~\cite{pomerleau1989alvinn}. Such direct control-based methods have advanced significantly in recent years thanks to deeper network architectures, more
\sm{informative} %
sensor inputs, and scalable learning methods~\cite{bojarski2016end,kendall2018learning,codevilla2018end,muller2018driving,cil,robust_bc,sim2real, philion2020lift}.
Recently, interpretable neural motion planners~\cite{nmp,dsdnet,abbas2020p3} provide an alternative that inherits the advantages of 
traditional pipelines and end-to-end approaches, by maintaining modularity and interpretability while enabling end-to-end learning.
This 
first began with joint perception and prediction \cite{faf, intentnet}, which neural planners extended to include planning.
Specifically, NMP~\cite{nmp} shared feature representations between multiple subtasks and predicted a cost volume to represent the quality of possible locations in planning. DSDNet~\cite{dsdnet} proposed an energy-based model to parameterize the joint distribution of \sm{the actors'} future trajectories. 
P3~\cite{abbas2020p3} developed a semantic occupancy representation and generated consistent ego-vehicle \sm{plans.}
Our work evaluates a wide range of \sm{autonomy}
systems, including modular and end-to-end interpretable ones.

\vspace{-2mm}
\paragraph{Safety-Critical Scenario Generation:}
There are three main components \sm{for generating}
safety-critical scenarios: a scenario parameterization space to optimize over, a search algorithm that identifies critical scenario parameters, and an evaluation setting to evaluate the system under test.
Previous works represent the action space of other agents either as a Frenet frame \cite{adv19evolutionary}, initial position and velocity \cite{ding2020multimodal, adv2020learn2collide}, a high-level graph-based route \cite{adv19bo}, or steering and acceleration \cite{adv19failure}. We choose to represent the behavior of actors as kinematic bicycle-model trajectories, allowing for physical feasibility and fine-grained behavior control.
There are many potential choices of the search algorithm 
used to identify 
scenarios that cause autonomy failure,
such as policy gradient \cite{adv2020lane, adv19failure, adv2020learn2collide},  Bayesian optimization \cite{adv19bo}, evolutionary algorithms  \cite{adv19evolutionary}, and variants of monte-carlo sampling \cite{ding2020multimodal, norden2019efficient, sinha2020neural}. 
\sm{We} build a general \sm{scenario generation} algorithm \sm{and}
benchmark a wide variety of black-box search algorithms, \sm{providing insight}
 into which search algorithms are 
 effective.
\cite{adv2020lane, adv19failure, adv2020learn2collide, adv19evolutionary} evaluate planners assuming groundtruth perception. %
\cite{adv19bo, norden2019efficient} use CARLA~\cite{carla}
to evaluate an image-based SDV planning algorithm. 
These works consider simplified planning modules or image-based systems that do not reflect
\sm{modern autonomy systems.} %
Moreover, they generate scenarios only for a handful of scene configurations.
In contrast, we present an end-to-end adversarial scenario generation system that takes into account failures of the full autonomy stack. 
Our approach scales to datasets with
diverse traffic patterns and map configurations.  We summarize the differences of our paradigm with prior works in Table~\ref{tab:prior_work}.

\begin{table}[t]
	\centering
	\resizebox{\linewidth}{!}{
		\begin{tabular}{c|cc|cc|cc}
			Method & 
			\rotatebox{90}{LiDAR} & \rotatebox{90}{Autonomy } & \rotatebox{90}{Traj.-based} & \rotatebox{90}{Real data} & \rotatebox{90}{ Scenarios} & \rotatebox{90}{\# Actors} \\ 
			\hline
			BOAdv~\cite{adv19bo} & & \checkmark & & & 2 & 2 \\ 
			Chen et al.~\cite{adv2020lane} & & & \checkmark  & & 1 & 3 \\
			Klischat et al.~\cite{adv19evolutionary} & & & & \checkmark  & 2 & 13 \\ 
			O’Kelly/Norden et al.~\cite{norden2019efficient,o2018scalable} & & \checkmark & \checkmark & & 1 & 5 \\ 
			\hline
		\textbf{AdvSim (ours)} & \boldcheckmark & \boldcheckmark & \boldcheckmark & \boldcheckmark & \textbf{4,000+} & 5 \\ 
		\end{tabular}
	}
\vspace{-2mm}
	\caption{\textbf{Comparison with prior works.} AdvSim produces safety-critical scenarios that are physically plausible and adversarial to full LiDAR-based autonomy systems in scale.} 
	\vspace{-4mm}
	\label{tab:prior_work}
\end{table}

\vspace{-2mm}
\paragraph{Physically Realizable Adversarial Examples:} 
Physical adversarial perturbations  expose real world threats for perception. Most previous works deal with threat models in the image space~\cite{advpatch,Athalye18,advtshirt,stopsign} by imposing physical constraints such as different view angles and distances~\cite{advpatch,Athalye18} and color distortion~\cite{stopsign,advtshirt}.
Recently, other works perturb
meshes~\cite{meshadv} or 
photo-realistic properties such as surface normals and lighting conditions~\cite{liu2018beyond,zeng2019adv}.
In 
self-driving, recent works focus on the robustness of LiDAR-based perception. %
Specifically, \cite{cao2020adv,james2020adv} 
generate objects that are either invisible or detrimental to object detectors and  \cite{lidarspoof,lidardefense} 
directly spoof the LiDAR points using laser devices. 
\cite{zhang2018camou,wu2020advcarla} craft adversarial vehicle textures in CARLA.
We instead generate
realistic adversarial maneuvers by other agents such that the scenario is safety critical for autonomy.

\begin{figure*}[t]
	\centering
	\includegraphics[width=\linewidth]{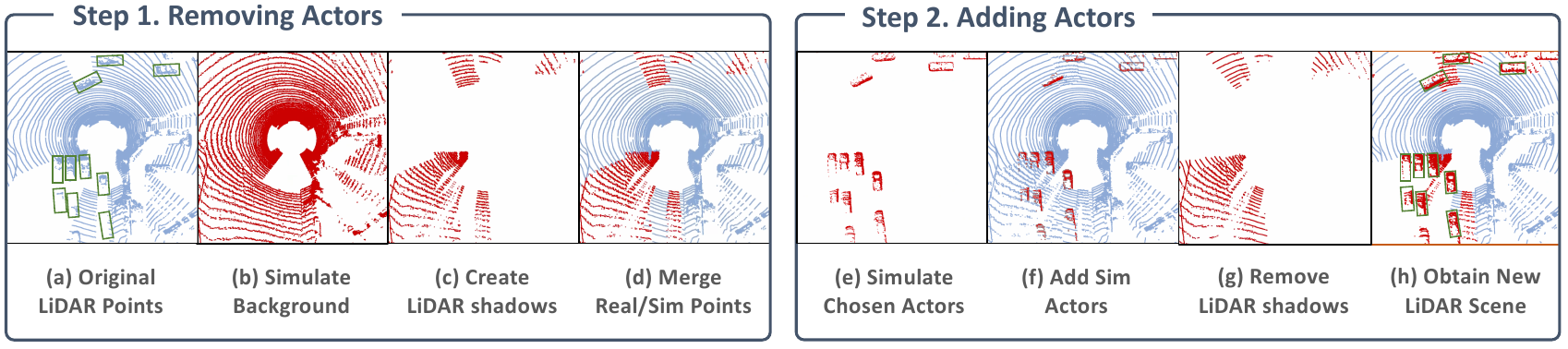} 
	\vspace{-0.25in}
	\caption{
\textbf{Realistic LiDAR simulation for scenario perturbations.} Given a scenario perturbation on the actors' motions, the previously recorded LiDAR data is modified to accurately reflect the updated scene configuration. We remove the original actor LiDAR observations and replace with simulated actor LiDAR observations at the perturbed locations, while ensuring sensor realism. The above example perturbs all actors left by 5 meters.
	}
	\label{fig:lidarsim}
\end{figure*}

\section{Generating Safety-Critical Scenarios}

Our objective is to generate realistic challenging scenarios that \sm{cause autonomy system failure.}
We frame our objective as a black box adversarial attack that exercises every component of the autonomy system, including object detection, motion forecasting and motion planning. 
As we search over the space of realistic perturbations in actor motions of an existing scenario, we must update the sensor data that the SDV observes and then evaluate the autonomy system.
Our approach, AdvSim, works as follows:
we first perturb the actors' motion trajectories in an existing scenario, and generate the sequence of LiDAR point clouds that reflect the change in actor locations. 
With the adjusted sensor data, we run the 
autonomy stack and get the planned SDV motion path. 
Finally, we evaluate the output path with a proposed adversarial objective and adjust the scenario perturbation to be more challenging.
An overview is shown in Fig.~\ref{fig:pipeline}.

In what follows, we first define the autonomy system and our attack formulation in  Sec~\ref{sec:problem_setup}.
We then describe how we parameterize the adversarial actors' behaviors (Sec~\ref{sec:model_adv_behaviors}) and conduct realistic LiDAR simulation to generate new LiDAR sweeps (Sec~\ref{sec:lidar_sim}).
Finally, we describe our adversarial objective
and the suite of black-box optimization algorithms we benchmark to generate worst-case behaviors in Sec~\ref{sec:searching_scenarios}.

\begin{algorithm}[t]
	\caption{Generating Adversarial Scenarios}\label{alg:scenario_gen}
	\begin{algorithmic}[1]
		\Require Sensory input $\mathbf{x}$, initial state $\mathbf{s}_0$ of the perturbed actor, adversarial objective $\mathcal{L}_{\mathrm{adv}}$, number of queries $N$. %
		\State Pick the perturbed actor  $\mathbf{v}_{\adv}$ heuristically
		\State Generate physically plausible trajectories set $\mathcal{T}_{\adv}$
		\State Initialize observation set $\mathcal{H} = \varnothing$
		\For {$k = 1, \dots, N$}
		\State Select $\boldsymbol{\delta}^{(k)}$ based on black-box algorithms and historical observations $\mathcal{H}$.
		\State $\tau_{\mathrm{adv}}^{(k)} = \Pi_{\tau \in \mathcal{T}_\adv}$ $\big[$\textsc{Bicycle} $\big(\mathbf{s}_0, \boldsymbol{\delta}^{(k)}\big)$$\big]$ \Comment{(Sec~\ref{sec:model_adv_behaviors})}
		\State $\mathbf{x}_\mathrm{adv}^{(k)} = f(\mathbf{x}, \tau_{\mathrm{adv}}^{(k)}, \tau_{\mathrm{sdv}})$ \Comment{(Sec~\ref{sec:lidar_sim})}
		\State Run the autonomy system and obtain the optimal SDV plan $\tau^{(k)}_0 = \tau^*_0(\mathbf{x}_{\adv}^{(k)})$
		\State Calculate the adversarial loss of the optimal plan:
		$\mathcal{L}^{(k)}_{\mathrm{adv}} = \mathcal{L}_{\mathrm{adv}} (\tau^{(k)}_0, \mathbf{x}^{(k)}_{\mathrm{adv}})$ \Comment{(Sec~\ref{sec:searching_scenarios})}
		\State Update observation set $\mathcal{H} = \mathcal{H} \cup \big\{(\tau_{\mathrm{adv}}^{(k)}, \mathcal{L}^{(k)}_{\mathrm{adv}})\big\}$
		\EndFor
		\State $\tau^*_{\mathrm{adv}} = \arg\max_{\tau^{(k)}_{\mathrm{adv}}, k\in [N]} \ \mathcal{L}^{(k)}_{\mathrm{adv}}$
	\end{algorithmic} 
\end{algorithm}

\subsection{Problem Setup}
\label{sec:problem_setup}
Let $\mathcal{V} = \{\mathbf v_0, \mathbf v_1, \dots \mathbf v_M\}$ be the set of vehicles that compose the scene, where $\mathbf v_0$ denotes the SDV, $M$ is the number of other vehicles.
The objective of a self-driving system is to find the best planned trajectory $\tau_0^*$ according to a cost function $\mathcal{C}$ that comfortably and safely maneuvers around the scene, given the available sensor data inputs $\mathbf x$:

\begin{equation}
\tau_0^* (\mathbf{x}) = \underset{\tau_0}{\arg\min}~\mathcal{C}(\tau_0, \mathbf{x})
\label{eq:cost}
\end{equation}
where $\tau_0$ is the SDV's planned  trajectory. 
As $\mathbf x$ consists of raw sensor data (i.e., LiDAR point clouds), High-Definition maps,  and other relevant information (e.g., previous SDV states,  traffic light states), this minimization represents the full autonomy system, not just the planning module. 

Our goal is to increase the risk of the self-driving car by perturbing the behaviors of other actors in a physically plausible manner for an existing traffic scenario.
Without loss of generality, we consider perturbing a single actor in the following discussion for brevity, but we apply AdvSim for multi-actor perturbations in
experiments.

We characterize the behavior of an adversary by the trajectory $\tau_{\mathrm {adv}}$ it will take in the future. %
As the perturbed actor's trajectory $\tau_{\adv}$ differs from its original behavior in the sensor data,
the vehicle position and
the occlusions it generates will change (see Fig. \ref{fig:lidarsim}). 
Therefore, we must simulate the new LiDAR data
given the adversary trajectory $\tau_{\adv}$ and SDV trajectory $\tau_{\mathrm{sdv}}$ to evaluate the system 
(Eq. \ref{eq:cost}). The generation of point clouds in the perturbed traffic scene is given as follows:
\begin{align}
\mathbf{x}_\mathrm{adv} &= f(\mathbf{x}, \tau_{\mathrm{adv}}, \tau_{\mathrm{sdv}})
\label{eq:query_func}
\end{align}
where $f(\cdot)$ denotes the realistic LiDAR simulation (Sec~\ref{sec:lidar_sim}) for perturbed input $\mathbf x_\mathrm{adv}$ given the adversary's trajectory and original sensor data sequence $\mathbf{x}$.

We then define an adversarial objective  $\mathcal{L}_{\mathrm{adv}}$  which we maximize to generate scenarios as follows
\begin{align}
\tau^*_{\mathrm{adv}} &= \underset{\tau_{\mathrm{adv}}}{\arg\max}~\mathcal{L}_{\mathrm{adv}}(\tau^*_0, \mathbf{x}_\mathrm{adv}),
\end{align}
where $\tau^*_0 = \tau^*_0(\mathbf{x}_{\mathrm{adv}})$ is the optimal SDV's planned trajectory under simulated scene $\mathbf{x}_\adv$. 
The design of the adversarial loss $\mathcal{L}_\adv$ is deferred to Sec~\ref{sec:searching_scenarios}.

\subsection{Modeling  Adversarial Behaviors}
\label{sec:model_adv_behaviors}
To produce physically feasible actor behaviors, we parameterize the trajectory $\tau_{\mathrm {adv}} = \{\mathbf s_t\}_{t=0}^T$ as a sequence of kinematic bicycle model states $\mathbf s_t = \{x_t, y_t, \theta_t, v_t, \kappa_t, a_t\}$ in the next $T$ timesteps. Here $(x,y)$ is the center position of the perturbed actor, $\theta$ is the heading, $v$ and $a$ are the forward velocity and acceleration, and $\kappa$ is the \sm{vehicle path's curvature.}
Candidate adversary trajectories can be generated by perturbing the change of curvature $\dot{\kappa}_t$ %
and acceleration values $a_t$ within set bounds at different timesteps, and using the kinematic bicycle model to compute the other states \cite{bicycle_model}.

Moreover, to enlarge the space of sampled adversarial behaviors, we also allow the perturbation of initial states ($x_0, y_0, \theta_0, v_0$) within set bounds.
In summary, the perturbation space can be depicted as
$$
\boldsymbol{\delta} = \left\{\Delta \mathbf{s}_0, \left(a_0, \dot{\kappa}_t|_{t=0} \right), \dots, \left(a_{T-1}, \dot{\kappa}_t|_{t={T-1}} \right) \right\}.
$$
To increase the perturbed trajectory's plausibility, we ensure it does not collide with other actors or the original expert trajectory of the SDV.
In practice, we do this by first performing rejection sampling to create a set of physically feasible trajectories $\mathcal{T}_{\adv}$ and then projecting the trajectory generated by $\boldsymbol{\delta}$ on to the physically feasible set, measured by $L_2$ distance.
Our search space is low-dimensional and conducive to query-based black box optimization, while still allowing for fine-grained actor motion control.

\subsection{Realistic LiDAR Simulation}
\label{sec:lidar_sim}
Given an initial traffic scenario and the corresponding adversarial perturbation to the actors' behaviors, we 
discuss how we modify the existing real LiDAR sweeps to reflect the perturbation.
We adopt the high-fidelity LiDARsim~\cite{siva2020lidarsim} simulator, which leverages real world data to generate realistic background meshes and dynamic object assets, and then applies physics-based raycasting and machine learning to generate realistic LiDAR point clouds. 
Given a modified scene configuration, we use LiDARsim to render a simulated point cloud, and then update the real LiDAR sweep with the modified regions.
We choose to update the sensor data for modified regions only rather than generating the full sweep to speed up the query function $f$ in Eq. \ref{eq:query_func}. Specifically, we cache the simulated background LiDAR (Fig.~\ref{fig:lidarsim} Step 1b) as the SDV trajectory is fixed during the actor perturbation.
\sm{The sensor perturbation}
is illustrated in Fig.~\ref{fig:lidarsim}. 

Modifying the LiDAR sensor data to reflect the scenario perturbation is non-trivial, as the LiDAR's sensing characteristics cause specific visibility artifacts that 
should exist in the generated scene to be realistic and physically accurate.
\sm{We perform} two main steps for sensor simulation for modified scenarios:
\textit{actor removal} (removing the existing benign actors' LiDAR point cloud and filling the LiDAR shadow created) and \textit{actor addition} (inserting the adversarial actors' LiDAR point cloud, while accounting for occlusion). 

\vspace{-2mm}
\paragraph{Removing Actors:} 
Given an original LiDAR point cloud (Fig.~\ref{fig:lidarsim}a), we first remove the points within the bounding boxes of perturbed actors
and simulate background points (Fig.~\ref{fig:lidarsim}b) using LiDARsim's background mesh.
We then convert the simulated and real LiDAR sweeps into a range image, 
allowing us to identify the specific rays missing in the real LiDAR sweep (Fig.~\ref{fig:lidarsim}c) 
that exist in the simulated LiDAR.
By taking the element-wise minimum ray distance between the range images, we can merge the LiDAR point clouds.
Fig.~\ref{fig:lidarsim}d shows the synthetic point clouds (red: simulated points; blue: original real points) after \textit{actor removal}.

\vspace{-2mm}
\paragraph{Adding Actors:} 
Once we have removed the selected actors from the LiDAR sweep, we 
update the LiDAR with the actors at their new locations. 
We first render the simulated LiDAR for the actors at their new locations using LiDARsim's vehicle asset bank (Fig.~\ref{fig:lidarsim}e).
Fig.~\ref{fig:lidarsim}f shows the real LiDAR point cloud with the added actors. 
However, when a LiDAR ray hits an object, the remaining path of the ray becomes occluded, creating a LiDAR shadow. 
Similar to the actor removal process, we create range images of the simulated and real LiDAR, and merge the LiDAR point clouds, thereby removing the LiDAR points of the now-occluded regions (Fig.~\ref{fig:lidarsim}g) and obtaining the final modified LiDAR sweep (Fig.~\ref{fig:lidarsim}h).
The generated scenes are realistic and match the desired perturbation in actors' motions (Fig.~\ref{fig:lidarsim}).

\setlength{\tabcolsep}{4.0pt}
\begin{table*}[t]
	\centering
	\resizebox{\textwidth}{!}{
		\begin{tabular}{llllllllllllll}
			\shline
			& \multicolumn{8}{c}{Planning Metrics $\downarrow$} & & \multicolumn{1}{c}{Perception $\uparrow$} & &  \multicolumn{2}{c}{Prediction Metrics} \\
			& \multicolumn{2}{c}{Collision (\%)} & & \multicolumn{2}{c}{L2 human} & & \multicolumn{2}{c}{Comfortable} & &
			\multicolumn{1}{c}{AP / F1 occ. (\%)} & & \multicolumn{2}{c}{L2 center / F1 occ. } \\
			\cline{2-3}\cline{5-6}\cline{8-9}\cline{11-11}\cline{13-14}
			& \multicolumn{1}{c}{up to $3 \mathrm{s}$} & \multicolumn{1}{c}{up to $5 \mathrm{s}$} & & \multicolumn{1}{c}{@3s} & \multicolumn{1}{c}{@5s} & & \multicolumn{1}{c}{Lat. ($m/s^2$)} & \multicolumn{1}{c}{Jerk ($m/s^3$) } & &
			& & \multicolumn{1}{c}{@3s} & \multicolumn{1}{c}{@5s} \\
			\rowcolor{grey}\multicolumn{14}{l}{\bt{IL: End-to-end Imitation Learning~\cite{bojarski2016end}}} \\
			Original & 5.7 & 20.4 & & 4.25 & 8.74 & & -- & -- & & -- & & -- & -- \\
			Adv. scenes & 8.9 \sbd{+3.2} & 25.2 \sbd{+4.8} & & 5.48 \sbd{+1.30} & 10.77 \sbd{+2.03} & & -- & -- & & -- & & -- & -- \\
			\rowcolor{grey}\multicolumn{9}{l}{\bt{NMP: Neural Motion Planner~\cite{nmp}}} & \multicolumn{2}{c}{AP (IoU = 0.7)} & \multicolumn{3}{c}{L2 center ($\downarrow$)}\\
			Original & 2.6 & 14.5 & & 3.30 & 9.31 & & 3.46 & 6.08 & & 81.7 & & 1.43 & 2.80 \\
			Adv. scenes & 14.2 \sbd{+11.6} & 42.2 \sbd{+27.7} & & 4.82 \sbd{+1.52} & 13.24 \sbd{+3.93} & & 3.59 \sbd{+0.13} & 7.28 \sbd{+0.97} & & 72.7 \sbd{-9.0} & & 1.63 \sbd{+0.20} & 3.21 \sbd{+0.41} \\
			\rowcolor{grey}\multicolumn{9}{l}{\bt{PLT: Jointly Learnable Behavior and Trajectory Planning~\cite{plt}}} & \multicolumn{2}{c}{AP (IoU = 0.7)} & \multicolumn{3}{c}{L2 center ($\downarrow$)}\\
			Original & 1.3 & 12.1 & & 1.74 & 5.10 & & 3.22 & 1.91 & & 83.5 & & 1.75 & 3.32 \\
			Adv. scenes & 5.5 \sbd{+4.2} & 31.6 \sbd{+19.5} & & 2.55 \sbd{+0.81} & 7.32 \sbd{+2.22} & & 3.51 \sbd{+0.29} & 2.38 \sbd{+0.47} & & 75.2 \sbd{-7.7} & & 1.96 \sbd{+0.21} & 3.75 \sbd{+0.43} \\
			\rowcolor{grey}\multicolumn{9}{l}{\bt{P3: Perceive, Predict, and Plan~\cite{abbas2020p3}}} & \multicolumn{2}{c}{F1 occ. (@0s)} & \multicolumn{3}{c}{F1 occ. (\%, $\uparrow$)}\\
			Original & 0.8 & 8.5 & & 1.58 & 4.74 & & 3.09 & 1.79 & & 64.9 & & 49.7 & 48.6 \\
			Adv. scenes & 3.9 \sbd{+2.9} & 32.0 \sbd{+23.5} & & 2.28 \sbd{+0.70} & 6.63 \sbd{+1.89} & & 3.18 \sbd{+0.09} & 2.18 \sbd{+0.30} & & 63.2 \sbd{-1.7} & & 47.5 \sbd{-2.2} & 45.9 \sbd{-2.7} \\
			\shline
		\end{tabular}
	}
	\vspace{-2mm}
	\caption{Evaluation of modern autonomy systems on original and AdvSim generated
scenarios. 
	\vspace{-2mm}
}
	\label{tab:eval_mp_models}
\end{table*}

\subsection{Adversarial Scenario Search}
\label{sec:searching_scenarios}
Since we aim for a general \sm{adversarial scenario generation} framework, 
we consider the autonomy system 
as a black box, where we 
access the evaluation scores through limited queries.
Our goal is to find the perturbation that maximizes \sm{the SDV's planned trajectory cost.}
In this section, we 
introduce the adversarial objective we optimize to produce worst-case scenarios \sm{and}
detail the search algorithms \sm{applied}.
We then summarize the 
AdvSim algorithm.

\vspace{-2mm}
\paragraph{Adversarial Objective:} 
To induce autonomy system failures, we propose a combination of three costs as our adversarial loss function. These costs are similar to those autonomy systems \cite{plt, ZieglerBDS14} attempt to minimize over in Eq. \ref{eq:cost}. We first include $l_{\mathrm{IL}}$, an imitation-learning based cost that encourages the SDV's output plan to deviate from the recorded human trajectory in the original scenario. We compute this as a smooth $\ell_{1}$ distance between output trajectory $\tau_0^*$ and the ground-truth human trajectory $\tau_h$ for the entire scenario horizon.
We also compute a cumulative collision (safety) cost $l_{\mathrm{col}}^{t}$  that encourages the perturbation to cause the SDV to collide with other actors in the scene.
Finally, we add other traffic-rule and comfort costs $c_{o}^t(\mathbf{x}_{\mathrm{adv}}, \tau_0^*)$ that encourages the output plan $\tau_0^*$ to have lane violations and be dangerous (i.e. high accelerations and jerk) at each timestep $t$.
The full adversarial loss is defined as:
\begin{align}
	\mathcal{L}_{\mathrm{adv}} &= l_{\mathrm{IL}} (\tau_{h}, \tau_0^*) + \sum_{t}l_{\mathrm{col}}^{t} (\mathbf x_{\mathrm{adv}}, \tau_0^*) + \sum_t  c_{o}^t(\mathbf{x}_{\mathrm{adv}}, \tau_0^*)  \nonumber
\end{align}

Our use of multiple different costs allows us to identify different types of autonomy system failures, such as unnatural trajectories, collisions, and hard braking.

\vspace{-2mm}
\paragraph{Search Algorithms:} 
AdvSim is a framework that can use any black-box search algorithm to identify autonomy system failures. The search algorithm attempts to find the safety critical scenarios by maximizing the adversarial objective $\mathcal L_{\adv}$ in Eq. \ref{eq:query_func}.
The search algorithm queries the autonomy system with a candidate perturbation $\tau_{\mathrm{adv}}$ to obtain a query pair $(\tau_{\mathrm{adv}}, \mathcal{L}_{\mathrm{adv}})$ and maintains a history $\mathcal H$ of past query pairs to generate the next candidate perturbation.
We study a wide variety of black-box search algorithms including (1) Bayesian optimization~\cite{gpucb,bayesopt_attack} (BO),  (2) genetic algorithms~\cite{genattack} (GA), (3) random search~\cite{simba} (RS) and (4) gradient estimation methods (NES~\cite{nes} and Bandit-TD~\cite{bandit_td}).
Specifically, BO maintains a surrogate model and select the next candidate based on the acquisition function and current model states. %
For GA algorithms, a group of candidate trajectories are evolved to maximize the objective and the best candidate is preserved at each iteration. %
For RS,
the perturbations sampled from a pre-defined orthonormal basis are added or subtracted to original input iteratively. 
Another branch of query-based black-box search algorithms 
estimate the gradient through the target model. Specifically, NES maximizes the expectation of the objective under one search distribution and Bandit-TD further leverages temporal information to improve the query efficiency. %

\vspace{-2mm}
\paragraph{Overall Adversarial Scenario Generation Algorithm:} 
We summarize our proposed AdvSim framework in Algorithm~\ref{alg:scenario_gen}. 
Given an initial traffic scene, we pick the actors to be perturbed using heuristics, such as the closest reachable actors, and then sample physically plausible trajectories $\mathcal{T}_{\adv}$ to ensure that our perturbations remain in this set. 
We then obtain the perturbation $\boldsymbol{\delta}^{(k)}$ at iteration $k$ based on historical observations $\mathcal{H}$ using a selected black-box search algorithm (L. 5).
We roll out the kinematics bicycle model states with initial state $\mathbf{s}_0$ and the perturbation $\boldsymbol{\delta}^{(k)}$, and project onto the feasible set $\mathcal{T}_{\adv}$ to obtain the adversarial trajectories for the perturbed actors (L. 6). 
After that, we update the sensor data accordingly (L. 7)
and evaluate the full autonomy system on generated scenarios to compute $\mathcal L_{\adv}$ (L. 8-9). 
Finally, after running the procedure for $N$ iterations, we obtain the adversarial behaviors of perturbed actors as well as corresponding simulated LiDAR data.

\section{Experiments}
We now showcase applying AdvSim to generate worst-case scenarios for several autonomy systems using a large scale self-driving dataset. In Sec. \ref{sec:exp_setup} we 
discuss the dataset and the autonomy systems under test, as well as how we evaluate the generated scenarios' effectiveness. In Sec. \ref{sec:exp_results} we 
\sm{analyze} how the \sm{AdvSim-}generated scenarios affect autonomy 
performance. 
\sm{We also show robust training on the generated scenarios improves autonomy systems.}%

\begin{figure*}[t]
	\centering
	\includegraphics[width=\linewidth]{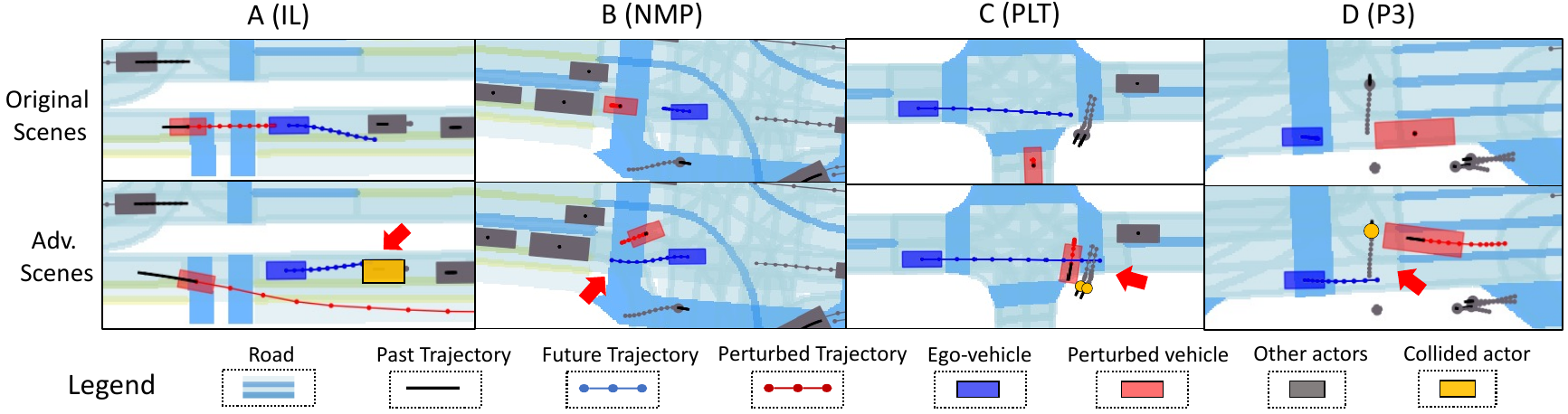}
	\vspace{-0.25in}
	\caption{
	    Visualization of autonomy system's output plan
		on original and corresponding adversarial scenes. \textbf{A:} 
		IL avoids the high-speed lane-changing vehicle behind %
		but collides
		with the front one. \textbf{B:} NMP collides with the \sm{merging vehicle.}
		\textbf{C:} PLT collides with one vehicle and two occluded pedestrians at crossroads. \textbf{D:} P3 collides with the \sm{crossing} pedestrian.
	}
	\vspace{-2mm}
	\label{fig:demo}
\end{figure*}
\begin{table}[t]
	\resizebox{\linewidth}{!}{
		\begin{tabular}{@{}c|c|c|c|c|c@{}}
			\toprule
			\multirow{2}{*}{Training} & \multirow{2}{*}{Testing} & Collision & L2 human & Lat. acc. & Jerk \\ 
			& & ($\%$ up to 5s) & @5s & $(m/s^3)$ & $(m/s^2)$ \\ \midrule
			\multirow{2}{*}{Standard
			} & Original & $8.5$ & $\mathbf{4.74}$ & $\mathbf{3.09}$ & $1.79$ \\
			& Adv. scenes & $32.0$ & $6.63$ & $\mathbf{3.18}$ & $2.18$ \\ \midrule
			\multirow{2}{*}{CL train} & Original & $\mathbf{7.0}$ & $4.88$ & $3.28$ & $1.81$ \\
			& Adv. scenes & $21.1$ & $5.98$ & $3.33$ & $2.08$ \\ \midrule	
			\multirow{2}{*}{Robust train} & Original & $7.3$ & $4.76$ & $3.29$ & $\mathbf{1.75}$ \\
			& Adv. scenes & $\mathbf{17.7}$ & $\mathbf{5.72}$ & $3.31$ & $\mathbf{2.04}$ \\ \bottomrule
		\end{tabular}
	}
	\centering
	\vspace{-2mm}
	\caption{Robust training P3 with augmented scenarios.}
	\vspace{-3mm}
	\label{tab:robust_train}
\end{table}

\subsection{Experimental Setup}
\label{sec:exp_setup}
\paragraph{Dataset}
We evaluate our approach on a 
self-driving dataset, \textbf{\textit{UrbanScenarios}},
which has 5,000 driving logs of 25 seconds each. Our dataset is collected across multiple cities in North America, and contains different types of map \sm{layouts} 
and varying traffic densities. 
We 
curate the dataset and select interesting candidate
scenarios to apply AdvSim on, where the SDV in the original scenario ``interacts" with other vehicles. 
Specifically, we sample 100 trajectories per SDV behavior (e.g., keep lane, lane change) in the SDV's Frenet frame~\cite{traj_sampler}
and calculate the \sm{trajectory collision rate}
with
other actors' motion paths. 
We select the
$6 \mathrm{s}$ scenario from each log
that has the largest collision rate. 
After data curation, we obtain 3953 train and 409 val scenarios.%

\vspace{-2mm}
\paragraph{Autonomy Systems}
We evaluate the effectiveness of the proposed framework on the following models: (a) \textbf{Imitation Learning (IL)}, where the future states of the SDV are predicted directly from the fused LiDAR and map features with $L_2$ loss; (b) \textbf{PLT}~\cite{plt}, a modular autonomy system where the detection and prediction are trained jointly with the backbone used in~\cite{abbas2020p3}, and the planning is
accomplished using a learnable combination of interpretable safety costs; (c, d) \textbf{NMP}~\cite{nmp} and \textbf{P3}~\cite{abbas2020p3}, two end-to-end interpretable motion planners. NMP predicts a cost-map directly from fused features with detection and prediction jointly trained as auxiliary tasks. P3 predicts a novel differentiable semantic occupancy representation used as safety-cost for planning. Please see \sm{supplementary}
for 
implementation details.

\vspace{-2mm}
\paragraph{Metrics:} 
In this paper, we focus on an open-loop scenario evaluation setting,
in which the evaluated autonomy system takes the past 1s LiDAR data as input
and outputs a 5s trajectory plan. We then unroll the 5s plan
and the other actors' trajectories for 5s and evaluate the autonomy system's performance during that time.
Following~\cite{dsdnet,plt,abbas2020p3}, we adopt standard planning metrics to measure the autonomy systems' performance on our adversarial scenarios, and compare how much more challenging they are relative to the original scenario set.
Specifically, \textbf{collision rate}
is the percentage of scenarios that cause the SDV to collide with another actor during a certain time frame (up to 3 or 5s).
\textbf{$\boldsymbol{L_2}$ distance to the human trajectory} represents how
well the model imitated the human driving.
\textbf{Jerk} and \textbf{lateral acceleration} indicate how comfortable the planned SDV trajectories are.
Additionally, all evaluated autonomy systems, except for \textbf{IL}, generate intermediate perception and prediction representations.
We therefore also report perception and motion forecasting metrics to see how the adversarial scenarios specifically impact these system components.
For NMP and PLT planners, for detection we report Average Precision (\textbf{AP})  
of bounding box
detections
at an Intersection of Union (IoU) of 0.7, and for motion forecasting we report
the $\boldsymbol{L_2}$ \textbf{prediction error} of predicted 
actor trajectories
at future timesteps. 
Similarly, for the P3 planner,
we adopt the F1-score of the occupancy prediction at different timesteps ($t = 0$ for detection and $t > 0$ for prediction).
All reported metrics are for vehicles in the ego-coordinate view range of $x \in [-72m, 72m], y \in [-40m, 40m]$.

\subsection{Experimental Results}
\label{sec:exp_results}

\paragraph{Evaluations on Modern Autonomy Systems:} 

We evaluate our AdvSim framework on the
autonomy systems 
in Table~\ref{tab:eval_mp_models}. 
Here, we only consider perturbing a single vehicle in the traffic scene and adopted Bayesian optimization (BO)~\cite{gpucb,bayesopt_attack} as the black-box search algorithm.
Experiments show that AdvSim is effective in generating challenging scenarios for different systems, leading to an average 
collision rate of $32.8\%$, over 200\% more compared to the original set. 
The adversarial scenarios also provide insights to how different autonomy systems compare.  
While NMP and PLT have similar drops in perception and motion forecasting, we observe that PLT is more robust to AdvSim scenarios than NMP on planning metrics, indicating the benefits of using a learning-based and hand-crafted cost function.
We show qualitative examples of adversarial scenarios for each autonomy system in
in Fig.~\ref{fig:demo}.
Unlike prior work, our generated scenarios cause the SDV to collide with other non-perturbed actors in the scene (Fig.~\ref{fig:demo}, A, C, D).

\begin{figure}
	\centering
	\includegraphics[width=\linewidth]{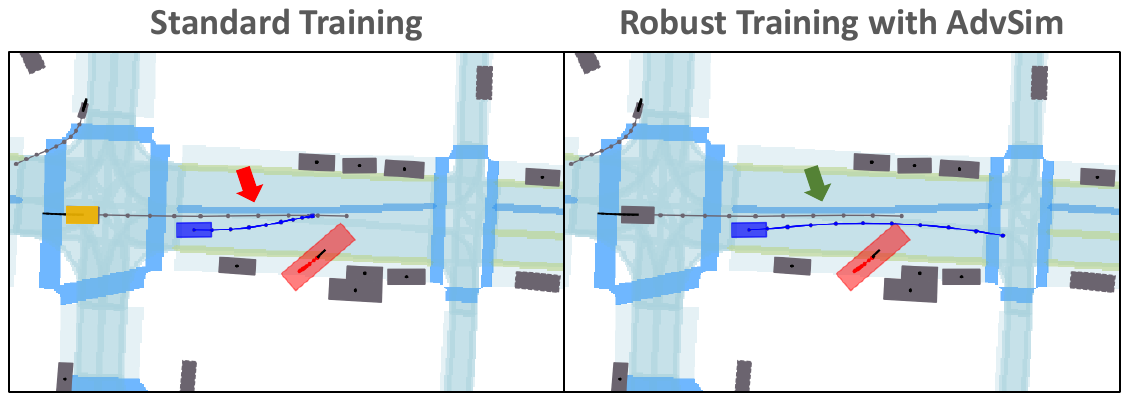}
	\vspace{-0.22in}
	\caption{Visualization of standard P3 and robust-trained P3 with AdvSim on one challenging scenario. Standard P3 changes lane to avoid the 
	reversing bus %
	yet is rear-ended from
	behind.
	 After robust training with AdvSim, the system 
	 bypasses the bus smoothly and returns to its original lane.%
	}
	\label{fig:robust_train}
\end{figure}

\vspace{-4mm}
\paragraph{Safer Planner with Challenging Scenarios:}
We investigate whether the robustness of the autonomy systems
can be improved with our generated
scenarios. 
We test several training schemes.
First, we propose a curriculum learning (CL)~\cite{BengioLCW09} baseline 
where we first train on standard examples till convergence (\textit{easy examples}), and then train on real challenging scenarios selected based on reachable actors (Sec~\ref{sec:exp_setup}) (\textit{hard examples}). 
Then, we propose a robust-training approach to leverage simulated worst-case scenarios.
Specifically, we use 
AdvSim to generate a large number of adversarial scenarios to augment the training data.
As discussed in Sec~\ref{sec:model_adv_behaviors}, the original expert
trajectory is still a valid planning solution to mimic in the
new scenario with respect to
collisions, as we impose 
constraints on the perturbation. 
This allows us to re-train autonomy systems with scenarios produced by AdvSim using the same expert trajectories as ground truth. 
Table~\ref{tab:robust_train} demonstrates that CL helps improve the performance on \textit{both} original and adversarial scenarios. 
Robust training with AdvSim-generated scenarios further improves performance across most planning metrics, highlighting the value of AdvSim scenarios for improving autonomy systems. 
In Fig.~\ref{fig:robust_train}, we show a qualitative example of an adversarial scenario for standard P3 and robust-trained P3 with AdvSim. Finally, we investigate the generalization of robust training by evaluating robustly trained PLT and P3 with safety-critical scenarios generated for other autonomy systems (see supplementary). 

\begin{table}[t]
	\centering
	\resizebox{0.9\linewidth}{!}{
		\begin{tabular}{@{}c|C{1.5cm}|C{1.5cm}|C{1.5cm}|C{1.5cm}@{}} \toprule
			\diagbox{Source}{Target} & IL & NMP & PLT & P3 \\ \midrule
			IL~\cite{bojarski2016end} & \underline{\textbf{25.2\%}} & 17.1\% & 9.5\% & 6.8\% \\
			NMP~\cite{nmp} & 24.9\% & \textbf{42.2\%} & 14.5\% & 14.0\% \\
			PLT~\cite{plt} & 21.2\% & 21.5\% & \textbf{31.6\%} & 13.1\% \\
			P3~\cite{abbas2020p3} & \textbf{26.4\%} & \underline{\textbf{24.2\%}} & \underline{\textbf{22.2\%}} & \textbf{32.0\%} \\ \bottomrule
		\end{tabular}
	}
	\vspace{-2mm}
	\caption{Transferability of generated safety-critical scenarios across different autonomy systems.}
	\vspace{-0.5mm}
	\label{tab:transferability}
	\vspace{-2mm}
\end{table}

\vspace{-3mm}
\paragraph{Transferability of Adversarial Scenarios:} 
We study the transferability of adversarial scenarios across different autonomy systems in Table~\ref{tab:transferability}, where \textit{Source} denotes the autonomy system used to identify failure scenarios and \textit{Target} denotes the autonomy system evaluated on these scenarios. 
We report the cumulative collision rate up to 5s. 
Results on additional metrics are \sm{in}
supplementary.
Table~\ref{tab:transferability} shows that generating adversarial scenarios with the same target autonomy system usually leads to the best performance. %
Scenarios simulated w.r.t the more robust P3 system (see original in Table~\ref{tab:eval_mp_models}) have stronger transferability. %

\begin{table}[t]
	\centering
	\resizebox{\linewidth}{!}{
		\begin{tabular}{c|c|c|cc|c}
			\toprule
			\multirow{2}{*}{\#Actors} & Collision &  L2 human & Lat. acc. & Jerk & F1 occ.\\
			& ($\%$, 5s) & @5s & $(m/s^2)$ & $(m/s^3)$ & ($\%$ @5s)\\
			\midrule 
			Original & $8.5$ & $4.74$ & $3.09$ & $1.79$ & $48.6$ \\ \midrule
			$m = 1$ & $32.0$ & $6.63$ & $3.18$ & $2.18$ & $45.9$   \\ 
			$m = 2$ & $\mathbf{33.5}$ & $7.21$ & $3.28$ & $2.26$ & $42.1$   \\  %
			$m = 3$ & $30.6$ & $7.10$ & $3.38$ & $2.24$ & $39.2$  \\ %
			$m = 4$ & $28.4$ & $7.36$ & $3.73$ & $2.46$ & $33.5$  \\ %
			$m = 5$ & $26.9$ & $\mathbf{7.86}$ & $\mathbf{4.62}$ & $\mathbf{2.62}$ & $\mathbf{30.5}$ \\ 
			\bottomrule
		\end{tabular}
	}
	\vspace{-2mm}
	\caption{Generating safety-critical scenarios for P3 with different number of perturbed actors $m$.}
	\vspace{-4mm}
	\label{tab:num_actors}
\end{table}

\vspace{-3mm}
\paragraph{Investigation on Attack Configurations:} 
We now study how the number of the perturbed actors and the search algorithm used \jk{affect} scenario generation.
In Table~\ref{tab:num_actors}, the planning metrics (L2 human, jerk and lateral acceleration) and the prediction metrics become worse as the number of perturbed actors $m$ increase. %
This indicates AdvSim generated more complicated traffic configurations when perturbing multiple actors simultaneously. 
However, we observed the collision rate decreases when $m \geq 3$. 
This may be because we sampled finite trajectories for each actor, and attacking \sm{with}multiple actors simultaneously increases the difficulty of finding physical plausible candidates to optimize over.  
We also benchmark a wide range of black-box algorithms in Table~\ref{tab:blackbox_algos}. %
See supplementary for the implementation details. 
We found that BO~\cite{bayesopt_attack} is most efficient,
 since the perturbation space is low dimensional and the cost measure is not smooth w.r.t the perturbation (thus harder for gradient-estimation based approaches). 

\begin{table}[t]
	\centering
	
	\resizebox{\linewidth}{!}{
		\begin{tabular}{c|c|c|cc|cc}
			\toprule
			\multirow{2}{*}{Algorithms} & Collision & L2 human & Lat. acc. & Jerk & \#Query. & GPU \\
			& ($\%$, 5s) & @5s & $(m/s^2)$ & $(m/s^3)$ & & Hour \\
			\midrule 
			Original & $8.5$ & $4.74$ & $3.09$ & $1.79$  & -- & -- \\
			GA~\cite{genattack} & $23.3$ &  $6.29$ & $3.13$ & $2.10$ &  $1600$ & $1.33$ \\ 
			NES~\cite{nes} & $19.1$ & $6.07$ & $3.17$  & $2.05$ & $400$ & $0.33$ \\ 
			Bandit-TD~\cite{bandit_td} & $14.6$ & $5.90$ & $3.12$ & $2.02$  & $100$ & $0.08$ \\
			RS~\cite{simba} & $14.7$ & $4.35$ & $3.14$ & $1.97$ & $100$ & $0.08$ \\
			BO~\cite{bayesopt_attack} & $\mathbf{28.5}$ & $\mathbf{6.63}$ & $\mathbf{3.18}$ & $\mathbf{2.18}$ & $\mathbf{75}$ & $\mathbf{0.06}$ \\ 
			\bottomrule
		\end{tabular}
	}
	\vspace{-2mm}
	\caption{Comparisons of different blackbox algorithms in scenario generation for P3.}
	\vspace{-2mm}
	\label{tab:blackbox_algos}
\end{table}

\vspace{-2mm}
\paragraph{Ablation Studies:}
We conduct ablation studies on proposed adversarial objective. 
As shown in Table~\ref{tab:ablation_loss}, imitation-learning based cost $l_\text{IL}$, cumulative collision cost $\sum_t \ell_{\mathrm{col}}^t$ and safety cost $\sum_t c_{s}^t $ are optimized for $L_2$ human,  collisions, and comfort planning metrics, respectively. The hybrid loss function ($\mathcal{M}_0$) \sm{can}
generate worst-case scenarios with respect to multiple metrics. 
If some planning metrics are particularly interesting in practice (e.g., collisions for testing), we could use a subset of \sm{the} proposed costs.
Unless otherwise stated, we adopt $\mathcal{M}_3$ in other experiments since the collisions are of key importance in \sm{evaluating autonomy systems.}
Furthermore, we compare $\mathcal{M}_0$ with other baseline adversarial objective in Table~\ref{tab:ablation_loss2}: (1) minimizing the closest distance to the ego-car~\cite{adv19bo}, (2) maximize the training cost proposed in~\cite{abbas2020p3}. Experiments show our design outperforms other baselines on all planning metrics.

\begin{table}
	\resizebox{\linewidth}{!}{
		\begin{tabular}{c|ccc|c|c|c|c}
			\toprule
			\multirow{2}{*}{\#ID} & IL & Collision & Safety & Collisions &  L2 human & Lat. acc. & Jerk \\
			& $\ell_{im}$ & $\sum_t \ell_{\mathrm{col}}^t$ & $\sum_t c_{s}^t $ & ($\%$ up to 5s) & @5s & $(m/s^3)$ & $(m/s^2)$ \\
			\midrule
			$\mathcal{M}_0$ & $\checkmark$ & $\checkmark$ & $\checkmark$ & $29.5$ & $6.12$ & $3.33$ & $2.17$    \\ %
			$\mathcal{M}_1$ & $\checkmark$ &  &  & $14.8$ & $\mathbf{7.06}$ & $3.22$ & $\mathbf{2.31}$    \\  %
			$\mathcal{M}_2$ &  & $\checkmark$ & & $30.6$ & $5.31$ & $3.01$ & $1.86$   \\ %
			$\mathcal{M}_3$ & $\checkmark$ & $\checkmark$ &  & $\mathbf{32.0}$ & $6.63$ & $3.18$ & $2.18$   \\ %
			$\mathcal{M}_4$ & & & $\checkmark$ & $9.0$ & $5.30$ & $\mathbf {3.52}$ & $2.08$   \\ %
			$\mathcal{M}_5$ & & $\checkmark$ & $\checkmark$ &  $30.6$ & $5.61$ & $3.44$ & $2.07$  \\ %
			\bottomrule
		\end{tabular}
	}
	\centering
	\vspace{-2mm}
	\caption{Ablation studies on adversarial objective design.}
	\vspace{-2mm}
	\label{tab:ablation_loss}
\end{table}

\begin{table}[t]
	\centering
	\resizebox{\linewidth}{!}{
		\begin{tabular}{c|c|c|c|c}
			\toprule
			& Collisions &  L2 human & Lat. acc. & Jerk \\
			& ($\%$ up to 5s) & @5s & $(m/s^3)$ & $(m/s^2)$ \\ \midrule
			Closest distance~\cite{adv19bo} & $13.5$  & $5.57$ & $3.16$ & $2.00$   \\ %
			Max-margin loss~\cite{abbas2020p3} & $18.3$  & $5.76$ & $3.02$ & $1.96$ \\ %
			$\mathcal{M}_3$ & $\mathbf{32.0}$ & $\mathbf{6.63}$ & $\mathbf{3.18}$ & $\mathbf{2.18}$   \\ %
			\bottomrule
		\end{tabular}
	}
	\vspace{-2mm}
	\caption{Comparison with other adversarial objective loss.}
	\vspace{-3mm}
	\label{tab:ablation_loss2}
	\centering
\end{table}

\section{Conclusion}
In this work we present a novel adversarial
framework to generate worst-case scenarios for modern autonomy systems.
Our approach identifies physically plausible failure cases that impose risks to full autonomy stack by simulating the sensor data based on the perturbed behaviors.
We demonstrate that AdvSim can generate failure cases at scale for a wide range of systems. %
More importantly, we leverage
these scenarios in training to further improve the robustness and safety of the autonomy system. 
We hope that leveraging this framework will allow for safer 
self-driving vehicles.

{\small
\bibliographystyle{ieee_fullname}
\bibliography{egbib}
}

\newpage 
\setcounter{section}{0}

\setcounter{section}{0}
\setcounter{figure}{0}
\makeatletter
\setcounter{table}{0}

\onecolumn
\appendix

\begin{center}
{\Large \bf Supplementary Material \\ AdvSim: Generating Safety-Critical Scenarios for Self-Driving Vehicles  \par}
\iftoggle{cvprrebuttal}{\vspace*{-22pt}}{\vspace*{24pt}}
{
	\large
	\lineskip .5em
	\begin{tabular}[t]{c}
	Jingkang Wang$^{1,2}$\quad Ava Pun$^{3}$\quad James Tu$^{1,2}$\quad Sivabalan Manivasagam$^{1,2}$\quad Abbas Sadat$^{2}$ \\ Sergio Casas$^{1,2}$\quad Mengye Ren$^{1,2}$\quad Raquel Urtasun$^{1,2}$ \\
{University of Toronto$^{1}$, Uber ATG$^{2}$, University of Waterloo$^{3}$} \\
{\tt\small {\{wangjk, manivasagam, sergio, mren, urtasun\}}@cs.toronto.edu}\ \ \\
\tt\small{a5pun@uwaterloo.ca} \ \ 
\tt\small{james.tu@mail.utoronto.ca} \ \ abbas.sadat@gmail.com
	\end{tabular}
	\par
}
\vskip .5em
\vspace*{12pt}
\end{center}

\begin{abstract}
	In this supplementary material, we first provide additional technical details on the AdvSim method and experiments.
	We provide details on our data curation (Sec~\ref{sec:data_curation}) for scenario generation, and how we generate physically realistic actor perturbations (Sec~\ref{sec:gen_realistic_trajs}). We also provide additional details for all the black box search algorithms used in the AdvSim framework (Sec~\ref{sec:black_box_algs}) and experimental details for robust training (Sec~\ref{sec:robust_training_supp}). 
	In the next section we show the results of generalization for robust training and additional metrics for the transfer learning experiment reported in the main paper (Sec \ref{sec:exps_supp}). 
	Then in Sec~\ref{sec:qualitative_exmaples} we provide additional qualitative examples for LiDAR simulation (Sec~\ref{sec:lidarsim_sup}), successful collision-inducing AdvSim scenarios (Sec~\ref{sec:scenarios_sup}), AdvSim scenarios that do not induce collision but still create challenges for the planning system (Sec~\ref{sec:non_col_scenes_sup}), as well as scenarios where multiple actors are perturbed simultaneously (Sec~\ref{sec:multi_actor}). 
\end{abstract}

\section{Implementation Details}
\subsection{Data Curation}
\label{sec:data_curation}
We evaluate our approach on a large-scale self-driving dataset, \textbf{\textit{UrbanScenarios}}, 
which covers 5,000 driving logs of 25 seconds each. However, it is difficult to simulate adversarial traffic scenarios if the SDV in the original scenario seldom ``interacts" with other vehicles (e.g., very few actors in the scene or actors are far away from ego-car). Therefore, we need to curate the dataset and select interesting sequences from which to generate safety-critical scenarios with AdvSim. 
Specifically, every 2s in the scenario sequence, we sample 100 trajectories per SDV behavior (e.g., keep lane, lane change) in the SDV's Frenet frame~\cite{traj_sampler}
and calculate the collision rate of these trajectories with
other actors' motion paths. 
We suppress false positive ``interactions'' by filtering the collisions with the vehicle directly in front of the SDV or with static vehicles.
Finally, we select the $6 \mathrm{s}$ scenario from each log that has the 
highest collision rate between trajectory samples and the actor future trajectories.
After data curation, we obtain 3953 train and 409 val scenarios in total. 
Note that the autonomy system is trained on all training logs of \textit{UrbanScenarios}.
Unless stated otherwise, we report the performance of the well-trained autonomy systems
on the adversarial scenarios generated on the validation set. 

\subsection{Generating Physical Plausible Trajectories}
\label{sec:gen_realistic_trajs}

\begin{algorithm}[htbp!]
	\caption{Generating Physical Plausible Trajectories}\label{alg:gen_trajs}
	\begin{algorithmic}[1]
		\Require Vehicle set $\mathcal{V}$, Sensory input $\mathbf{x}$, total number of sampled behaviors $N_\text{sample}$, minimum number of feasible trajectories $N_{\min}$.
		\State $\mathcal{T}_\adv = \varnothing$
		\While{$|\mathcal{T}_\adv| > N_{\min}$}
		\If {$|\mathcal{V}| = 0$} \Return \textsc{None}
		\Else {\ $\mathcal{T}_\adv = \varnothing$}
		\EndIf
		\State Select the closest actor ($\mathbf{v}_{\mathrm{adv}} \in \mathcal{V}$) w.r.t SDV.
		\State Sample $N_\text{sample}$ trajectories $\mathcal{T}_{\adv}$ for actor $\mathbf{v}_{\mathrm{adv}}$.
		\State Prune physically implausible trajectories in $\mathcal{T}_\adv$.
		\State Remove chosen actor in set: $\mathcal{V} = \mathcal{V} \setminus \{\mathbf{v}_\adv\}$.
		\EndWhile
		\State \Return $\mathcal{T}_\adv$
	\end{algorithmic}
\end{algorithm}

To ensure our trajectory perturbation for an actor is physically feasible, we create a feasible trajectory set for the actor. We can then project the trajectory perturbation onto the feasible set to get the output perturbation.
First, we select the actors in the region of interest and select the closest $m$ actors to ego-car over the planning horizon.
In our experiments, we randomly sample $N_\text{sample} = 2e^{5}$ trajectories for each perturbed actor, and then prune physically implausible ones to obtain the feasible candidate set. 
Specifically, we remove trajectories that collide with other actors or the expert trajectory of the SDV over the entire 6s horizon (1s observation + 5s planning). 
We also remove trajectories that drive off the road, according to the HD map.
We perturb actors that have at least
$N_{\min} = 100$ trajectories in the feasible set to guarantee that there are enough candidates for searching.
The whole procedure of actor selection and generating physical plausible candidate trajectories is depicted in Algorithm~\ref{alg:gen_trajs}.

To produce physical plausible perturbations, we set realistic bounds for the kinematic bicycle model in Table~\ref{tab:constraints_bicycle}. Moreover, we normalize the perturbation $\boldsymbol{\delta}$ such that it becomes $\ell_\infty$-norm ball with the perturbation magnitude $\epsilon = 1.0$

\begin{table}[htbp!]
	\centering
	\resizebox{0.42\linewidth}{!}{
		\begin{tabular}{@{}c|c@{}}
			\toprule
			\textbf{Physical Constraints} & \textbf{Value} \\ \midrule
			Curvature $\kappa$ ($m^{-1}$)& $[-0.2, 0.2]$ \\ 
			Change of curvature $d\kappa/dt$ ($m^{-1}t^{-1}$) & $0.05$ \\
			Acceleration $a$ ($m/s^2$) & $[-2.0, 2.0]$ \\
			Lateral acceleration ($m/s^2$)& $[-3.0, 3.0]$			
			\\ 
			Velocity ($m/s$) & $15$ \\ \midrule
			Initial position perturbation $\Delta x_0, \Delta y_0$ & $[-5.0, 5.0]$\\ 
			Initial velocity perturbation $\Delta v_0$ & $[-5.0, 5.0]$\\ 
			Initial heading perturbation $\Delta \theta_0$ & $[-\pi / 4, \pi / 4]$\\ 
			\bottomrule
		\end{tabular}
	}
	\vspace{-2mm}
	\caption{Physical constraints on the kinematic bicycle model and the perturbation for initial state $\mathbf{s}_0$.}
	\label{tab:constraints_bicycle}
\end{table}

\subsection{Black box Search Algorithms}
\label{sec:black_box_algs}
We investigate a wide range of black box search algorithms for AdvSim. In what follows, we introduce each algorithm and provide the implementation details.

\paragraph{Bayesian Optimization (BO)~\cite{gpucb,bayesopt_attack}:}
BO maintains a surrogate model and select the next query candidate based on the current model states (historical observations) and the acquisition function that strikes a balance of exploration and exploitation. Specifically, we adopt standard Gaussian process (GP) model with upper confidence bound (UCB) as the acquisition function. The exploration multiplier is $\beta = 3$.
The total query budget for BayesOpt attack is set as 75 and the first 20 queries are used for the initialization of GP model. Since the attack space is low-dimensional thus we do not use the additive GP models and auto subspace search techniques used in ~\cite{bayesopt_attack}.

\paragraph{Genetic Algorithm (GA)~\cite{ManTK96,genattack}:}
GA is a population-based algorithm for black-box optimization. Specifically,
a group of candidate trajectories are evolved to maximize the fitness score (adversarial objective) and the best candidate is preserved at each iteration. Then new candidates are generated by sampling new behaviors from historical candidates, with sampling probability proportional to the fitness score. We adopt the adaptive parameter scaling~\cite{genattack} to lessen the sensitivity of genetic algorithms to hyperparameter values:
\begin{align}
	p &= \max(p_{\min}, p_0 \times \gamma^g), \\
	z &= \max(z_{\min}, z_0 \times \gamma^g),
\end{align}
where $p$ and $z$ are mutation rate and magnitude respectively, $\gamma$ and $g$ are the exponential decay rate and number of plateaus. $(\cdot)_{\min}$ and $(\cdot)_{0}$ represents the minimum value and initial value for the mutation rate $p$ or mutation range $z$. The detailed hyperparamter setting is provided in Table~\ref{tab:ga_hparams}. 

\begin{table}[htbp!]
	\centering
	\resizebox{0.38\linewidth}{!}{
		\begin{tabular}{@{}c|c@{}}
			\toprule
			\textbf{Hyperparamter} & \textbf{Value} \\ \midrule
			Population size  & $32$ \\
			Number of generations  & $50$ \\
			Number of maximum plateaus & $25$ \\ \midrule
			Initial mutation range & $\mathcal{U}{(1.0, 1.0)}$ \\
			Minimum mutation range & $\mathcal{U}{(-0.3, 0.3)}$ \\
			Initial mutation probability $p_0$ & $0.5$ \\
			Minimum mutation probability $p_{\min}$ & $0.1$ \\
			Exponential decay rate $\gamma$ & $0.9$ \\ \bottomrule
		\end{tabular}
	}
	\vspace{-2mm}
	\caption{Hyperparamter setting for genetic algorithm.} %
	\label{tab:ga_hparams}
\end{table}

\paragraph{Random Search (RS)~\cite{simba,square_attack}:}

Recent study shows that adding or subtracting random perturbations iteratively is efficient in producing strong threats to neural networks. Specifically, one random perturbation is sampled from a pre-defined orthonormal basis and added to the original input at each iteration. If it enhances the adversarial objective, we accept the sampled perturbation, otherwise, we update the current input with an inverse perturbation.
We simply adopt the Cartesian basis and do not utilize dimension reduction techniques~\cite{simba,square_attack} since the perturbation space is low-dimensional in our setting.
The perturbation magnitude at each iteration is set as $\epsilon^{(k)} = 0.25$ and we run for 100 iterations in total.

\paragraph{Gradient Estimation Approaches~\cite{nes,bandit_td}:} Finally, we study another branch of query-based optimization algorithms that estimate the gradient through the target model.
Specifically, natural
evolution strategies (NES)~\cite{nes} maximizes the expectation of the objective under one search distribution, and gradient estimation with bandits (Bandit-TD)~\cite{bandit_td} further leverages time-dependent priors to improve the query efficiency. With estimated gradients, we leverage the projected gradient descent (PGD) algorithm~\cite{pgd} to optimize the perturbation. We adopt the official PyTorch implementations\footnote{\url{https://github.com/MadryLab/blackbox-bandits}}. The hyperparameters for the NES and bandits approaches are presented in Table~\ref{tab:nes_bandit}. We refer the reader to~\cite{bandit_td} for more details of the NES and Bandit-TD algorithms.

\begin{table}[htbp!]
	\centering
	\begin{subtable}[h]{0.25\linewidth}
		\centering
		\resizebox{\textwidth}{!}{
			\begin{tabular}{@{}c|c@{}}
				\toprule
				\textbf{Hyperparamter} & \textbf{Value} \\ \midrule
				Samples per step  & $10$ \\
				Learning rate  & $0.25$ \\
				Finite different probe & $0.5$ \\ 
				Number of iterations & $20$	\\ \bottomrule
			\end{tabular}
		}
		\caption{NES}
	\end{subtable}
	\hspace{5mm}
	\begin{subtable}[h]{0.28\linewidth}
		\centering
		\resizebox{\textwidth}{!}{
			\begin{tabular}{@{}c|c@{}}
				\toprule
				\textbf{Hyperparamter} & \textbf{Value} \\ \midrule
				Prior learning rate  & $1.0$ \\
				Perturbation learning rate & $0.25$ \\
				Finite different probe & $0.5$ \\ 
				Bandit exploration & $1.0$	\\ \bottomrule
			\end{tabular}
		}
		\caption{Bandit-TD}
	\end{subtable}
	\caption{Hyperparamters for (a) NES and (b) Bandit-TD.}	\label{tab:nes_bandit}
\end{table}

\subsection{Robust Training}
\label{sec:robust_training_supp}
For robust training, we first generate safety-critical scenarios with AdvSim on both training and validation set. Then we first
train the P3 model~\cite{abbas2020p3} for $80k$ iterations till convergence. Then we retrain this model with selected challenging scenarios (CL baseline) or generated adversarial scenarios (robust train) for $10k$ iterations with a learning rate of $2e^{-4}$. We adopted the early stopping technique during the re-training process.

\section{Additional Experiments}
\label{sec:exps_supp}

\subsection{Generalization of Robust Training}

We investigate the generalization of robust training in Table~\ref{tab:robust_train_transfer} by evaluating  %
robustly trained PLT and P3 with safety-critical scenarios generated for other autonomy systems. 
The cumulative collision rate up to 5s is reported. %
Robustly trained systems  outperform standard systems consistently on adversarial scenarios generated from different autonomy systems.

\begin{table}[htbp!]
	\centering
	\resizebox{0.53\linewidth}{!}{
		\begin{tabular}{@{}c|c|C{1.2cm}|C{1.2cm}|C{1.2cm}|C{1.2cm}@{}}
			\toprule
			\multicolumn{2}{c|}{\diagbox{Training}{Testing}} & IL & NMP & PLT & P3 \\ \midrule
			\multirow{2}{*}{PLT} & Standard & 9.5 & 14.5 & 31.6 & 22.2 \\
			& Robust train & \textbf{7.3} & \textbf{12.9} & \textbf{14.3} & \textbf{15.1} \\ \midrule
			\multirow{2}{*}{P3} & Standard & 6.8 & 14.0 & 13.1 & 32.0 \\
			& Robust train & \textbf{5.3} & \textbf{11.6} & \textbf{10.8} & \textbf{17.7} \\ \bottomrule
		\end{tabular}
	}
	\caption{Generalization of robustly trained systems (PLT and P3) across safety-critical scenarios generated with respect to different autonomy systems.}
	\label{tab:robust_train_transfer}
\end{table}

\subsection{Supplementary Metrics for Table 3}
Similar to Table 3, we report the L2 prediction error (L2 human) at 5s to investigate the transferability of adversarial scenarios across different autonomy systems.
Table~\ref{tab:transferability_sup} demonstrates that generating adversarial scenarios with the same target autonomy system usually leads to the best performance. 

\begin{table}[htbp!]
	\centering
	\resizebox{0.53\linewidth}{!}{
		\begin{tabular}{@{}c|C{1.5cm}|C{1.5cm}|C{1.5cm}|C{1.5cm}@{}} \toprule
			\diagbox{Source}{Target} & IL & NMP & PLT & P3 \\ \midrule
			IL~\cite{bojarski2016end} & \textbf{10.77} & 10.52 & 5.20 & 5.15 \\
			NMP~\cite{nmp} & 8.41 & \textbf{13.24} & 5.21 & 5.11 \\
			PLT~\cite{plt} & 8.00 & 9.91 & \textbf{7.32} & 5.19 \\
			P3~\cite{abbas2020p3} & 8.37 & 10.76 & 5.23 & \textbf{6.63} \\ \bottomrule
		\end{tabular}
	}
	\vspace{-2mm}
	\caption{Transferability of generated safety-critical scenarios across different autonomy systems (L2 human).}
	\label{tab:transferability_sup}
\end{table}

\section{Qualitative Examples}
\label{sec:qualitative_exmaples}
In this section, we first provide the qualitative examples for simulated LiDAR sweeps (Fig.~\ref{fig:lidarsim_supp}) during the scenario generation procedure. We then present more visualizations of AdvSim Scenarios and corresponding motion forecasting predictions (Fig.~\ref{fig:adv_scenes_supp}). We then show scenarios generated by AdvSim that are more challenging for the SDV, but do necesarilly cause collision due to the robustness of the autonomy system under test (Fig.~\ref{fig:failure_case}).
Finally, we provide the visualizations of adversarial scenarios where multiple vehicles are perturbed (Fig.~\ref{fig:multi_actor}).

\subsection{Realistic LiDAR Simulation for AdvSim}
\label{sec:lidarsim_sup}

\begin{figure}[htbp!]
	\centering
	\includegraphics[width=\linewidth]{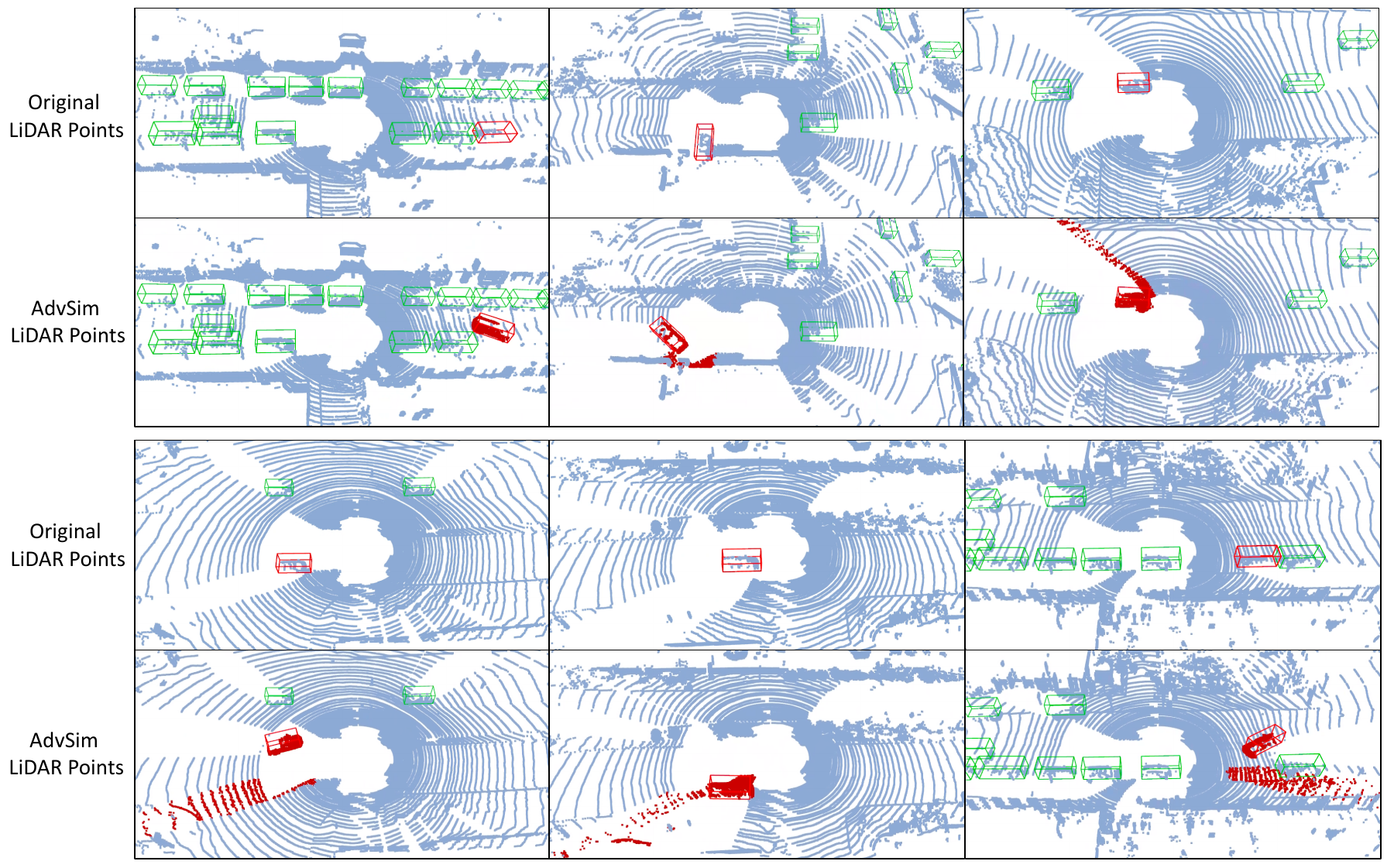}
	\caption{Comparison of original and simulated LiDAR sweeps. The \textcolor{blue}{blue} and \textcolor{red}{red} points represent the \textcolor{blue}{original} and \textcolor{red}{simulated} LiDAR data respectively. We also visualize the bounding boxes of the perturbed actor (\textcolor{red}{red}) and the other actors (\textcolor{ForestGreen}{green} for reference.
		\label{fig:lidarsim_supp}
	}
\end{figure}

We show more examples of the original and corresponding simulated LiDAR sweeps after actor perturbation in Fig.~\ref{fig:lidarsim_supp}. The simulated points for the perturbed actor and the background are red. The original LiDAR points are blue. We also highlight the bounding boxes of the perturbed actor in both original and AdvSim traffic scenes. Fig.~\ref{fig:lidarsim_supp} demonstrates that our generated LiDAR scenes are realistic and match the desired perturbation in actors' motions well.

\subsection{More Visualizations of AdvSim Scenarios}
\label{sec:scenarios_sup}
\begin{figure}[htbp!]
	\centering
	\includegraphics[width=\linewidth]{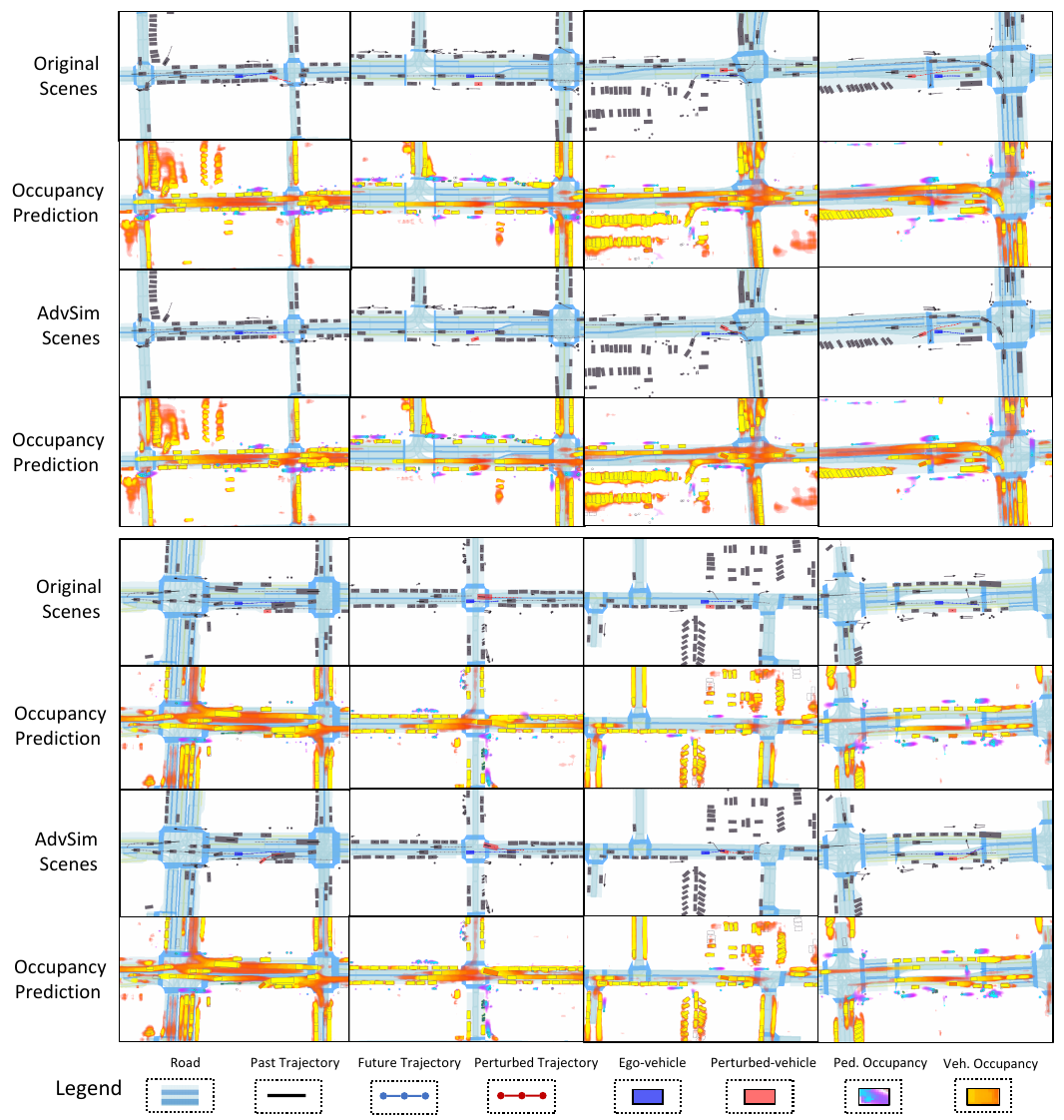}
	\caption{More visualizations of the P3 autonomy system's output plans on original and corresponding adversarial scenes. We show eight AdvSim scenarios where the SDV collides with the other actors (e.g., pedestrian, perturbed vehicle or other vehicles). Please zoom-in for better viewing of the scenario}
	\label{fig:adv_scenes_supp}
\end{figure}

In Fig.~\ref{fig:adv_scenes_supp}, we provide more AdvSim scenarios where the P3 autonomy system's plans cause collision with other actors in the planning horizon. To understand how the SDV behaves under these newly generated scenarios, we further visualize the occupancy prediction for motion forecasting from in the planning horizon (blue-purple for pedestrian, yellow-red for vehicle).

\subsection{Non-collision Scenarios Generated by AdvSim}
\label{sec:non_col_scenes_sup}
\begin{figure}[htbp!]
	\centering
	\includegraphics[width=\linewidth]{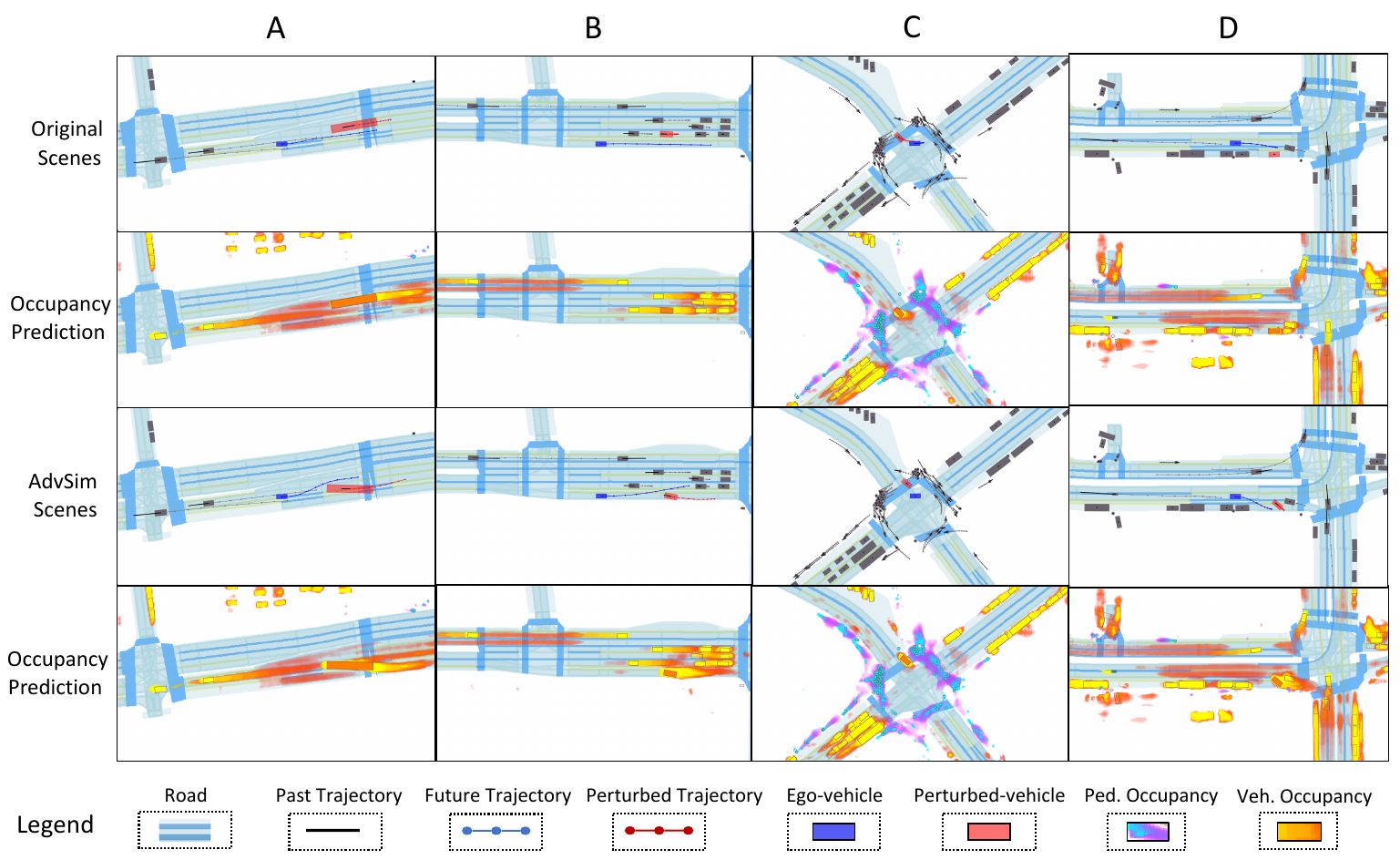}
	\caption{Visualizations of the adversarial scenarios where the P3 autonomy system performs reasonably and AdvSim is unable to  cause collision. \textbf{A:} 
		The SDV bypasses the perturbed bus smoothly by changing to another lane. \textbf{B:} Although the front lane is occupied by the perturbed vehicle, the SDV is capable of avoiding collisions by changing to another lane smoothly. \textbf{C:} The SDV keeps stationary as there are many pedestrians walking in the crosswalks. \textbf{D:} The SDV pulls over smoothly since the lane is occupied by the perturbed actor.}
	\label{fig:failure_case}
\end{figure}

In Fig.~\ref{fig:failure_case}, we present some qualitative examples of adversarial scenarios generated by AdvSim where the P3 autonomy system produces reasonable plans to react. Specifically, these examples are where AdvSim is unable to generate a scenario perturbation that causes the P3 system to collide with other actors. However, these AdvSim scenarios are still potentially challenging since the SDV needs to produce plans with larger costs. These scenarios can provide additional insights and are still valuable for testing and analysis.

\subsection{Perturbing Multiple Actors with AdvSim}
\label{sec:multi_actor}
\begin{figure}[htbp!]
	\centering
	\includegraphics[width=\linewidth]{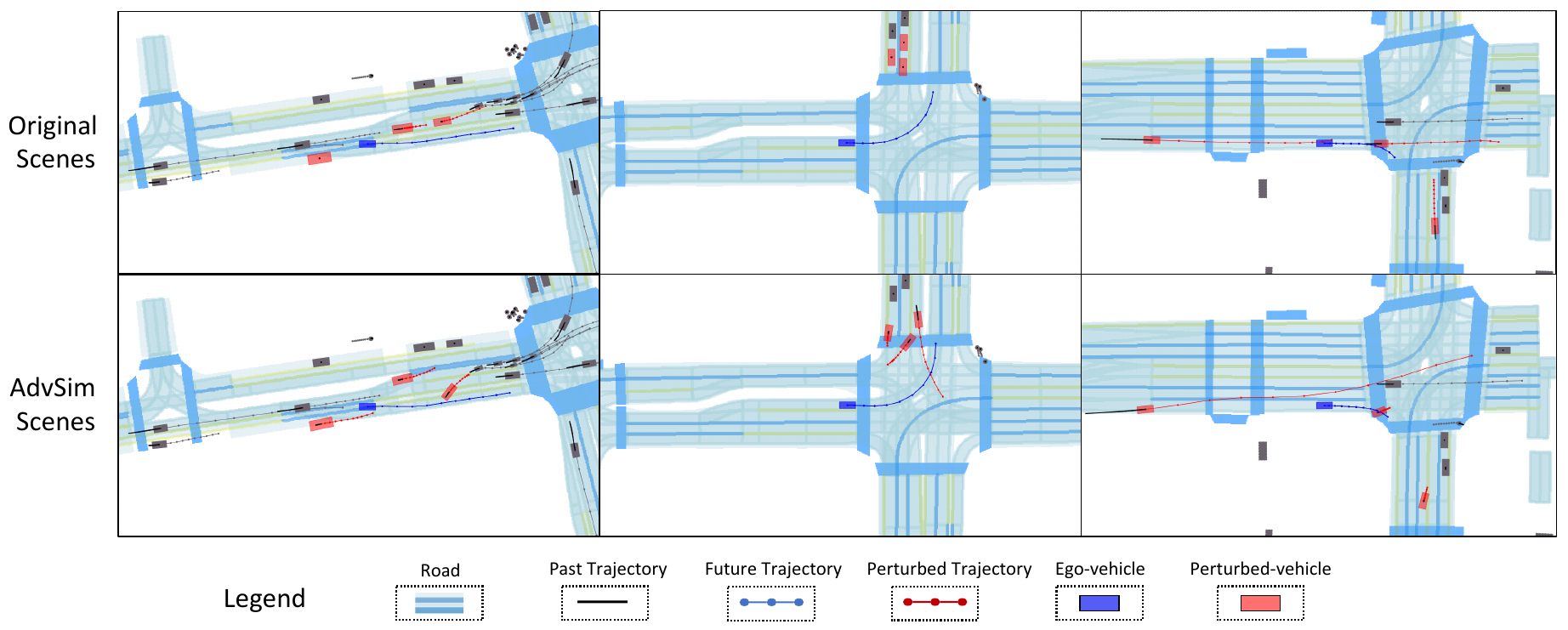}
	\caption{Visualizations of the adversarial scenarios where multiple actors are perturbed by AdvSim.}
	\label{fig:multi_actor}
\end{figure}

Finally, we present qualitative examples for multi-agent perturbations evaluated in  Table 4. As shown in Fig.~\ref{fig:multi_actor}, %
our proposed AdvSim could produce physically plausible behaviors for multiple actors and generate scenarios with difficulty to self-driving systems.

\end{document}